\documentclass[nohyperref]{article}

\usepackage{microtype}
\usepackage{graphicx}
\usepackage{subfigure}
\usepackage{booktabs} 

\usepackage{hyperref}

\usepackage[preprint]{icml2023}

\usepackage{amsmath}
\usepackage{amssymb}
\usepackage{mathtools}
\usepackage{amsthm}
\usepackage{bm}
\usepackage{bbm}
\usepackage{multirow}

\DeclareMathOperator*{\argmax}{arg\,max}

\usepackage[capitalize,noabbrev]{cleveref}

\theoremstyle{plain}

\theoremstyle{definition}

\theoremstyle{remark}

\usepackage[textsize=tiny]{todonotes}

\icmltitlerunning{Long-tailed Classification from a Bayesian-decision-theory Perspective}

\begin{document}

\twocolumn[
\icmltitle{Long-tailed Classification from a Bayesian-decision-theory Perspective}



\icmlsetsymbol{equal}{*}

\begin{icmlauthorlist}
\icmlauthor{Bolian Li}{purdue}
\icmlauthor{Ruqi Zhang}{purdue}
\end{icmlauthorlist}

\icmlaffiliation{purdue}{Department of Computer Science, Purdue University, West Lafayette, IN, USA
}

\icmlcorrespondingauthor{Ruqi Zhang}{ruqiz@purdue.edu}

\icmlkeywords{Long-tailed classification, Bayesian Decision Theory}

\vskip 0.3in
]



\printAffiliationsAndNotice{}  

\begin{abstract}
Long-tailed classification poses a challenge due to its heavy imbalance in class probabilities and tail-sensitivity risks with asymmetric misprediction costs. Recent attempts have used re-balancing loss and ensemble methods, but they are largely heuristic and depend heavily on empirical results, lacking theoretical explanation. Furthermore, existing methods overlook the decision loss, which characterizes different costs associated with tailed classes. This paper presents a general and principled framework from a Bayesian-decision-theory perspective, which unifies existing techniques including re-balancing and ensemble methods, and provides theoretical justifications for their effectiveness. From this perspective, we derive a novel objective based on the integrated risk and a Bayesian deep-ensemble approach to improve the accuracy of all classes, especially the ``tail". Besides, our framework allows for task-adaptive decision loss which provides provably optimal decisions in varying task scenarios, along with the capability to quantify uncertainty. Finally, We conduct comprehensive experiments, including standard classification, tail-sensitive classification with a new False Head Rate metric, calibration, and ablation studies. Our framework significantly improves the current SOTA even on large-scale real-world datasets like ImageNet.
\end{abstract}

\section{Introduction}
Machine learning methods usually assume that training and testing data are both i.i.d. (independent and identically distributed) sampled from the same data distribution. However, this is not always true for real-world scenarios~\cite{hand2006classifier}. One example is \emph{long-tailed classification}~\cite{reed2001pareto,lin2014microsoft,van2017devil,krishna2017visual,liu2019large,wang2020long,li2022trustworthy}, where the training data is biased towards a few ``head" classes, while the ``tailed" classes have fewer samples, resulting in a ``long-tailed" distribution of class probabilities. The long-tailed problem is mainly due to the process of collecting data, which is unavoidably biased. Conventional models trained on long-tailed data often report significant performance drops compared with the results obtained on balanced training data~\cite{wang2022partial}. Besides, for some real-world applications, the risk of classifying tailed samples as head (which is a common type of mistakes) is obviously more severe than that of classifying head samples as tail (which is less common)~\cite{sengupta2016frontal,rahman2021nurse,yang2022proco}. The significant performance drop and the ``tail-sensitivity risk" limit the application of ML models to long-tailed classification.

Existing works usually re-balance the loss function to promote the accuracy of tail classes, which typically re-weights the loss function by a factor $1/f(n_y)$, where $f(\cdot)$ is an increasing function and $n_y$ refers to the number of samples in the class $y$~\cite{lin2017focal,cao2019learning,cui2019class,wu2020distribution}. Their re-weighting strategy compensates for the lack of training samples in tailed classes, but suffers from sub-optimal head class accuracies. Other attempts on ensemble models try to reduce the model variance to promote the head and tail accuracies at the same time~\cite{wang2020long,li2022trustworthy}. Despite the effectiveness of existing works, they suffer from significant limitations: \textbf{\romannumeral1)} their algorithm design is largely based on empirical results without adequate theoretical explanation; \textbf{\romannumeral2)} they do not consider the decision loss, which represents the application-related risks (e.g., the tail-sensitivity risk) in the real world and thus their models are not applicable to tasks with different metrics other than standard classification task with accuracy; \textbf{\romannumeral3)} most methods do not quantify uncertainty in their predictions, which reduces their reliability. These limitations undermine the potential of existing works for real-world long-tailed data.

In this paper, we propose a unified framework for long-tailed classification, rooted in \emph{Bayesian Decision Theory}~\cite{berger2013statistical,robert2007bayesian}. Our framework unifies existing methods and provides theoretical justifications for their effectiveness, including re-balancing loss and ensemble methods, which have been shown to achieve promising results. To derive our framework, we first introduce a new objective based on the \emph{integrated risk} which unifies three crucial components in long-tailed problems: data distribution, decision loss, and posterior inference. To minimize this objective, we then derive a tractable lower bound based on variational EM~\cite{lacoste2011approximate} and approximate the posterior by a particle-based ensemble model~\cite{liu2016stein,d2021repulsive}. Furthermore, we design two kinds of \emph{utility functions} for the standard and tail-sensitive classifications respectively, which enables real-world applications with tail-sensitivity risks. Finally, we conduct comprehensive experiments on three long-tailed datasets under various evaluation metrics to demonstrate the superiority of our method in general settings. We summarize our contributions as follows:
\begin{itemize}
    \item \emph{Long-tailed Bayesian Decision} (LBD) is the first to formulate long-tailed classification under Bayesian Decision Theory, providing a fresh perspective and unifying three key components of long-tailed problems (data distribution, posterior inference and decision making) within a single principled framework.
    \item We propose a new objective based on \emph{integrated risk}, which exploits variational approximation, utility functions and an efficient particle-based BNN model. It significantly enhances the flexibility (task-adaptive utility functions) and reliability (uncertainty quantification) of our framework.
    \item For real-world applications, we take the decision loss into account, extending our method to more realistic long-tailed problems where the risk of wrong predictions varies and depends on the type of classes (e.g., head or tail). We also design a new metric (False Head Rate) to evaluate this kind of risk accordingly.
    \item We conduct comprehensive experiments including generalization accuracy, uncertainty estimation, and a newly designed False Head Rate (FHR) to show the effectiveness of our method on various tasks. In particular, our method outstrips the SOTA in large-scale real-world datasets like ImageNet on all metrics.
\end{itemize}

\section{Related Works}
\paragraph{Long-tailed Classification.} To overcome long-tailed class distributions, over-sampling~\cite{han2005borderline} uses generated data to compensate the tailed classes, under-sampling~\cite{liu2008exploratory} splits the imbalanced dataset into multiple balanced subsets, and data augmentation~\cite{chu2020feature,kim2020m2m,liu2020deep} introduces random noise to promote model's robustness. Recent advances focus on improving training loss functions and model architectures. For example, re-weighting methods~\cite{cao2019learning,cui2019class,lin2017focal,menon2020long,wu2020distribution,mahajan2018exploring} adjust the loss function by class probabilities in the training data, OLTR~\cite{liu2019large} transfers the knowledge learned from head classes to the learning of tailed classes, LFME~\cite{xiang2020learning} uses multiple teacher models to learn relatively balanced subsets of training data, RIDE~\cite{wang2020long} develops a multi-expert framework to promote the overall performances with ensemble model, and TLC~\cite{li2022trustworthy} exploits the evidential uncertainty to optimize the multi-expert framework. Besides, SRepr~\cite{namdecoupled} explores Gaussian noise in stochastic weight averaging to obtain stochastic representation, and SADE~\cite{zhangself} considers the case of non-uniform testing distributions in long-tailed problems. These methods are largely designed based on empirical heuristics, and thus their performances are not explainable and guaranteed. In contrast, our method is rooted in Bayesian principle and decision theory, inheriting their theoretical guarantees and explanation.

\paragraph{Bayesian Decision Theory.} Bayesian Decision Theory is introduced in \citet{robert2007bayesian,berger2013statistical}. It provides a bridge which connects posterior inference, decision, and data distribution. Loss-calibrated EM~\cite{lacoste2011approximate} exploits the posterior risk~\cite{schervish2012theory} to simultaneously consider inference and decision. \citet{cobb2018loss} further extends this method using dropout-based Bayesian neural networks. Loss-EP~\cite{morais2022loss} applies the technique of loss calibration in expectation propagation. Post-hoc Loss-calibration~\cite{vadera2021post} develop an inference-agnostic way to learn the high-utility decisions. These methods all use the notion of utility to represent their prior knowledge about the application-related risks, and exploit the posterior risk. While they show great advantages in some applications, none of them consider the data distribution, which prevents their applications on long-tailed data. Our method overcomes this limitation by introducing the integrated risk, which unifies data distributions, inference, and decision-making.

\paragraph{Ensemble and Particle Optimization.} Ensemble models combine several individual deep models to obtain better generalization performances~\cite{lakshminarayanan2017simple,ganaie2021ensemble}, which is inspired by the observation that multiple i.i.d. initializations are less likely to generate averagely ``bad" models~\cite{dietterich2000ensemble}. Ensemble models can also be used to approximate the posterior with the technique of \emph{particle optimization}, which is first studied in Stein variational gradient descent (SVGD,~\citet{liu2016stein}) and then explored by \citet{liu2019understanding,korba2020non,d2020annealed}. \citet{liu2017stein} analyzes SVGD in a gradient-flow perspective. \citet{wang2018function} performs the particle optimization directly in the function space. \citet{chen2018unified,liu2018accelerated} put the particle optimization in the 2-Wasserstein space. \citet{d2021repulsive} implements particles by introducing a repulsive force in the gradient flow. Instead of directly modeling the gradient flow, our framework optimizes the particles through stochastic gradient descent (SGD,~\citet{bottou1998online}), with repulsive force induced by the integrated risk objective. Compared to existing particle optimization, our method is easy and cheap to implement, which is especially beneficial for large deep models.

\section{Background}
\subsection{Long-tailed Distribution}
Long-tailed distributed data is a special case of \emph{dataset shift}~\cite{quinonero2008dataset}, in which the common assumption is violated that the training and testing data follow the same distribution~\cite{moreno2012unifying}. For the long-tailed scenario studied in this paper, the training data $\mathcal{D}_{train}$ is distributed in a descending manner over categories in terms of class probability:
\begin{equation}
    p(\bm{x}_1,y_1=k_1)\geq p(\bm{x_2},y_2=k_2),~~\text{if}~k_1\leq k_2
\end{equation}
for all $(\bm{x}_1,y_1),(\bm{x}_2,y_2)\in\mathcal{D}_{train}$, while the testing data $\mathcal{D}_{test}$ is assumed to be distributed uniformly over categories:
\begin{equation}
    p(\bm{x}_1,y_1=k_1)=p(\bm{x_2},y_2=k_2)
\end{equation}
for all $(\bm{x}_1,y_1),(\bm{x}_2,y_2)\in\mathcal{D}_{test}$. One important feature of long-tailed distribution is that both training and testing data are semantically identical\footnote{In contrast to the open-set scenario~\cite{geng2020recent}, where additional classes may cause testing data to be semantically irrelevant to training data.}, and the only difference lies in class probabilities.

\subsection{Bayesian Decision Theory}
Bayesian Decision Theory is a general statistical approach and can be applied to the task of pattern classification~\cite{berger2013statistical,robert2007bayesian}. In standard Bayesian inference, for a training dataset $\mathcal{D}_{train}=\{(\bm{x}_i,y_i)\}_{i=1}^N$ and a $\theta$-parameterized model, we have the likelihood $\prod_i p(y_i|\bm{x}_i,\bm{\theta})$ and prior $p(\bm{\theta})$. The Bayesian approach tries to estimate the posterior $p(\bm{\theta}|\mathcal{D}_{train})=p(\bm{\theta})\prod_i p(y_i|\bm{x}_i,\bm{\theta})/\prod_ip(y_i|\bm{x}_i)$. For test data $x^*$, we obtain the predictive distribution $p(y^*|\bm{x}^*,\mathcal{D}_{train})=\int_\theta p(y^*|\bm{x}^*,\bm{\theta})p(\bm{\theta}|\mathcal{D}_{train})d\bm{\theta}$ by averaging over all possible models weighted by their posterior probabilities. The Bayesian Decision Theory further considers the utility of making different decisions and the data distribution, which bridges the posterior inference, data distribution, and decision-making in a unified manner. For example, the \emph{posterior risk} is defined by the decision losses averaged over the posterior, and the \emph{integrated risk} further considers the data distribution. Bayesian Decision Theory has theoretical guarantees on the results and is provable to provide a desirable decision. For example, \citet{robert2007bayesian} illustrates two kinds of optimalities, \emph{minimaxity} and \emph{admissibility}, of the model that minimizes the integrated risk. Therefore, models following Bayesian Decision Theory are expected to have smaller risks than models trained in other ways. We in this paper simultaneously consider posterior, decision loss, and long-tailed data distribution in a unified Bayesian framework.

\section{Long-tailed Bayesian Decision}
For conventional frameworks in long-tailed classification, one crucial challenge is that: \textbf{inference} (how to infer model parameters), \textbf{decision} (model's actions in the presence of application-related risks), and \textbf{data distribution} (long-tailed distribution) are independent from each other in the training phase~\cite{lacoste2011approximate}. To the best of our knowledge, none of previous methods can simultaneously consider these three aspects. In order to address this drawback, we introduce the \emph{integrated risk} from Bayesian Decision Theory, which is computed over the posterior $p(\bm{\theta}|\mathcal{D})$ and the data distribution $p(\bm{x},y)$:
\begin{equation}
    R(d):=\mathbb{E}_{(\bm{x}_i,y_i)\sim p(\bm{x},y)}\mathbb{E}_{\bm{\theta}\sim p(\bm{\theta}|\mathcal{D})}l(\bm{\theta},d(\bm{x}_i)),\label{eq:ir}
\end{equation}
where $l(\bm{\theta},d(\bm{x}_i))$ is the loss of making decision $d(\bm{x}_i)$ for $\bm{x}_i$ when the environment is $\bm{\theta}$ (model's parameters). The decision estimator $d$ that minimizes the integrated risk is proved to give the optimal decisions in terms of the decision loss~\cite{robert2007bayesian}.

In order to exploit Eq.~\ref{eq:ir} as the objective, we need to determine the posterior and the optimal decision at the same time, which is notoriously hard because they are dependent on each other. Inspired by the Expectation-Maximization (EM) algorithm~\cite{dempster1977maximum,lacoste2011approximate}, which alternately conducts the integration and optimization steps, we propose a long-tailed version of variational EM algorithm to alternately update a variational distribution and a classification decision on long-tailed data. 
To use EM, we convert the minimization problem to a maximization problem. Specifically, we define the \emph{decision gain} $g(\bm{\theta},d(\bm{x}_i))\propto -l(\bm{\theta},d(\bm{x}_i))$ to be:
\begin{equation}
g(\bm{\theta},d(\bm{x}_i)):=\prod_{y'}p(y'|\bm{x}_i,\bm{\theta})^{u(y',d(\bm{x}_i))}
\end{equation}
to represent what we gain from making decision $d(\bm{x}_i)$ given the environment $\bm{\theta}$ (see Appendix~\ref{sec:decision_gain} for more details). Here, $u(y',d)$ is a fixed utility function that gives the utility of making decision $d$ when the true label is $y'$. Then our goal becomes maximizing the \emph{integrated gain}:
\begin{equation}
    G(d):=\mathbb{E}_{(\bm{x}_i,y_i)\sim p(\bm{x},y)}\mathbb{E}_{\bm{\theta}\sim p(\bm{\theta}|\mathcal{D})}g(\bm{\theta},d(\bm{x}_i)).\label{eq:ig}
\end{equation}
We will work with this objective and discuss the details of Long-tailed Bayesian Decision in the following sections.

\subsection{Task-adaptive Utility Functions}
\begin{figure}[t]
    \centering
    \includegraphics[width=8cm]{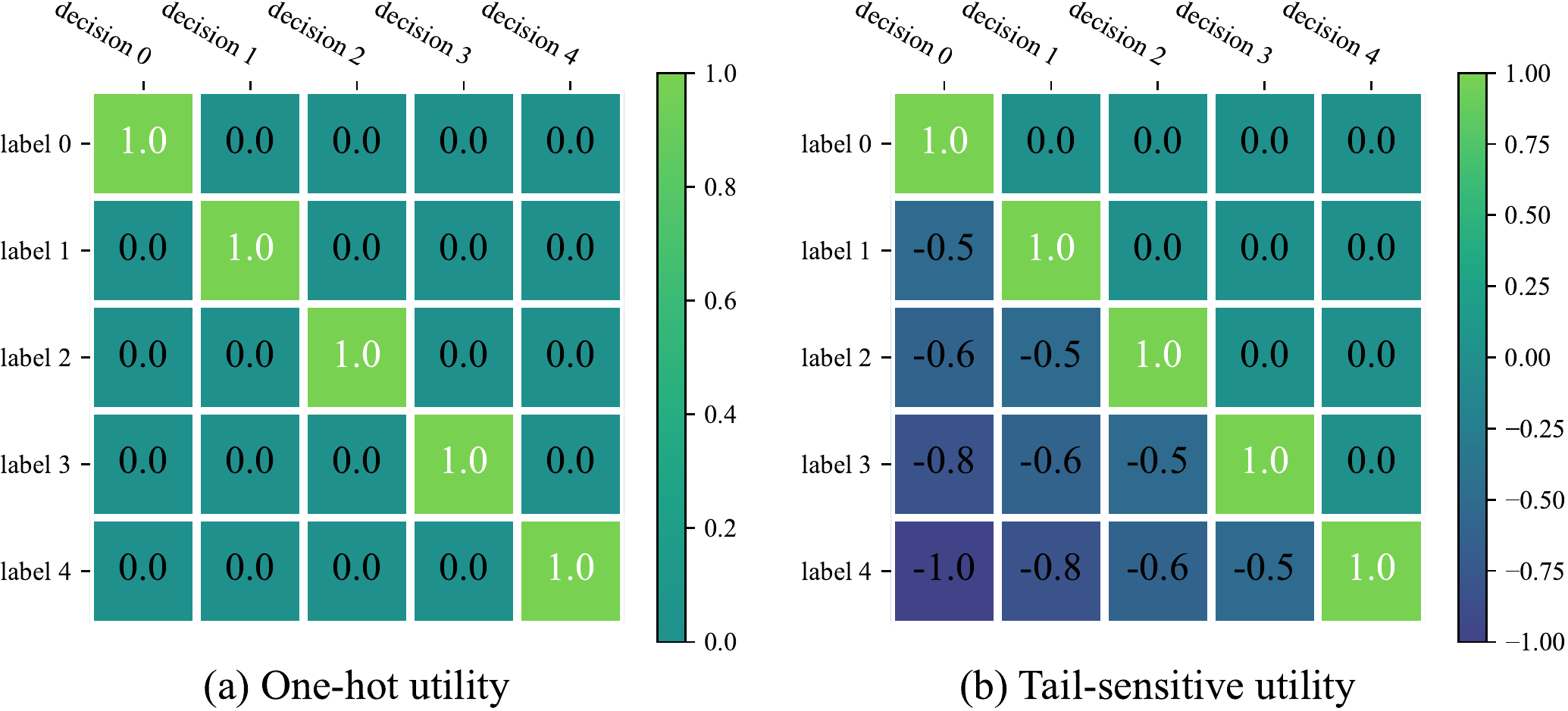}
    \caption{Two examples of utility matrices, designed for (a) standard and (b) tail-sensitive classifications respectively.}
    \label{fig:u}
\end{figure}
We first discuss the design of the utility function: $u(y,d)$, where $y$ is the ground truth (true label) and $d$ is the decision (predicted label). The utility function defines the gain of making different decisions and can encode our preference for specific metrics in various tasks. The utility function is a standard component in Decision Theory and its design has been comprehensively studied in the literature. For example, Chapter 2.2 of \citet{robert2007bayesian} guarantees the existence of utility functions with rational decision-makers. Generally, the values of the utility function over all class labels are stored in a form of utility matrix $\bm{U}$, where $U_{ij}=u(y=i,d=j)$.

In a standard classification setting, the overall accuracy is the most decisive metric in evaluation. It only matters whether the decision is consistent with the ground truth (i.e., $y=d$). Therefore, as shown in Fig.~\ref{fig:u}(a), a simple one-hot utility can be defined by $u(y,d)=\mathbbm{1}\{y=d\}$, which is corresponding to the standard accuracy metric. 

In modern applications of long-tailed classification, the semantic importance of ``tailed" data often implies more penalty in the circumstance of predicting tailed samples as head~\cite{sengupta2016frontal,rahman2021nurse,yang2022proco}. Besides, the lack of training samples in tailed classes has been empirically proved to be the bottleneck of classification performance~\cite{li2022trustworthy}. Therefore, the ratio of false head samples in evaluation would reflect the potential of a model in real-world applications~\footnote{The quantitative form will be discussed in Section~\ref{subsec:fhr}.}. To this end, a tail-sensitive utility can be defined by adding an extra penalty on those false head samples, as shown in Fig.~\ref{fig:u}(b). The tail-sensitive utility encourages the model to predict any uncertain sample as tail rather than head, while not affecting the predictions of the true class when the model is confident.

\subsection{Inference Step}
Due to the discrepancy between training (long-tailed) and testing (uniform) data distributions, we propose to compute the integrated gain with the posterior of \emph{testing} data $p(\bm{\theta}|\mathcal{D}_{test})$ to target at evaluation, where $\mathcal{D}_{test}=\{(\bm{x}_i,y_i)\}_{i=1}^N$ with $(\bm{x}_i,y_i)\sim p_{test}(\bm{x},y)$. To infer the posterior $p(\bm{\theta}|\mathcal{D}_{test})$, we use the variational method where a variational distribution $q(\bm{\theta})$ is introduced to the lower bound of the integrated gain in Eq.~\ref{eq:ig}:
\begin{equation}
\begin{aligned}
    &\log G(d)=\log\mathbb{E}_{(\bm{x}_i,y_i)\sim p_{test}(\bm{x},y)}\mathbb{E}_{\bm{\theta}\sim p(\bm{\theta}|\mathcal{D}_{test})}g(\bm{\theta},d(\bm{x}_i))\\
    &\geq\mathbb{E}_{(\bm{x}_i,y_i)\sim p_{train}(\bm{x},y)}\mathbb{E}_{\bm{\theta}\sim q(\bm{\theta})}\frac{p_{test}(\bm{x}_i,y_i)}{p_{train}(\bm{x}_i,y_i)}\cdot \\
    &~~~~~~~~\left[\log{p(y_i|\bm{x}_i,\bm{\theta})}+\sum_{y'}u(y',d)\log{p(y'|\bm{x}_i,\bm{\theta})}\right] \\
    &~~~~-KL(q(\bm{\theta})||p(\bm{\theta}))+C \\
    &:= L(q,d),\label{eq:obj}
\end{aligned}
\end{equation}

where $(\bm{x}_i,y_i)$ is training data but we forcefully compute its probability on testing distribution, and $C$ is a constant. Eq.~\ref{eq:obj} is proved in Appendix~\ref{appendix:e} and the main idea is to apply Jensen’s inequality~\cite{jensen1906fonctions}. The lower bound $L(q,d)$ is our training objective and it provides a cross-entropy-like way to update the variational distribution $q(\bm{\theta})$, and most importantly, converts the data distribution from $p_{test}(\bm{x},y)$ (uniform) to $p_{train}(\bm{x},y)$ (long-tailed) by the technique of importance sampling~\cite{kloek1978bayesian} to make the computation during training possible\footnote{Thanks to the fact that training and testing data are semantically identical.}. 

Moreover, the variational distribution $q(\bm{\theta})$ is guaranteed to be an approximation of the posterior $p(\bm{\theta}|\mathcal{D}_{test})$, because Eq.~\ref{eq:obj} contains Bayesian inference on the posterior of testing data. To support this, we look into the KL divergence between $q(\bm{\theta})$ and $p(\bm{\theta}|\mathcal{D}_{test})$:
\begin{equation}
\begin{aligned}
    KL&(q(\bm{\theta})||p(\bm{\theta}|\mathcal{D}_{test})) \\
    &=-\mathbb{E}_{(\bm{x}_i,y_i)\sim p_{train}(\bm{x},y)}\mathbb{E}_{\bm{\theta}\sim q(\bm{\theta})}\frac{p_{test}(\bm{x}_i,y_i)}{p_{train}(\bm{x}_i,y_i)} \\
    &~~~~~~~~\cdot\log{p(y_i|\bm{x}_i,\bm{\theta})}+KL(q(\bm{\theta})||p(\bm{\theta}))-C,\label{eq:bi}
\end{aligned}
\end{equation}
where $C$ is a constant. Eq.~\ref{eq:bi} is proved in Appendix~\ref{appendix:kl}. Comparing Eq.~\ref{eq:obj} and Eq.~\ref{eq:bi}, it is clear that part of the objective at inference step is Bayesian inference on the posterior of testing data, with $q(\bm{\theta})$ approaching $p(\bm{\theta}|\mathcal{D}_{test})$.\footnote{In fact, $q(\bm{\theta})$ approximates the gain-calibrated posterior: $\tilde{p}(\bm{\theta}|\mathcal{D}_{test})\propto p(\bm{\theta}|\mathcal{D}_{test})\prod_{y'}p(y'|\bm{x}_i,\bm{\theta})^{u(y',d(\bm{x}_i))}$.}

In summary, the objective $L(q,d)$ in Eq.~\ref{eq:obj} enables the framework to simultaneously consider inference, decision (utility), and data distribution ($p_{test}(\bm{x},y)/p_{train}(\bm{x},y)$, which will be further discussed in Section~\ref{subsec:dis}). It provides a principled way to optimize the integrated gain in Eq.~\ref{eq:ig}.

\subsection{Decision Step}
To optimize $L(q,d)$ w.r.t the decision $d$, one way is to select the decision $d^\star$ that maximizes the gain respectively for each input $\bm{x}_i$ given the current variational distribution:
\begin{equation}
    d^\star=\argmax_{d}\mathbb{E}_{\bm{\theta}\sim q(\bm{\theta})}\sum_{y'}u(y',d)\log{p(y'|\bm{x}_i,\bm{\theta})}.\label{eq:d}
\end{equation}
Notably, for symmetric utility functions (e.g., one-hot utility), Eq.~\ref{eq:d} can be further simplified: $d^\star=\argmax_{d}\mathbb{E}_{\bm{\theta}\sim q(\bm{\theta})}\log{p(d|\bm{x},\bm{\theta})}$, which is equivalent to the maximum of the predictive distribution. 

However, during training, we essentially know that the optimal decisions for training data are their true labels. Therefore, we can utilize this knowledge and simply set $d(x_i)=y_i$. We can also view this as selecting the optimal decisions under a well-estimated $q(\bm{\theta})$ in Eq.~\ref{eq:d} instead of the current distribution, since we expect $d^\star$ approach the true labels as $q(\bm{\theta})$ keeps updating. Then the objective can be further simplified to be:
\begin{equation}
\begin{aligned}
&L(q,d)=L(q) \\
&=\mathbb{E}_{(\bm{x}_i,y_i)\sim p_{train}(\bm{x},y)}\mathbb{E}_{\bm{\theta}\sim q(\bm{\theta})}\frac{p_{test}(\bm{x}_i,y_i)}{p_{train}(\bm{x}_i,y_i)}\cdot \\
    &~~~~~~~~\left[\log{p(y_i|\bm{x}_i,\bm{\theta})}+\sum_{y'}u(y',y_i)\log{p(y'|\bm{x}_i,\bm{\theta})}\right] \\
    &~~~~-KL(q(\bm{\theta})||p(\bm{\theta}))+C.\label{eq:final_obj}
\end{aligned}
\end{equation}

During testing, we use Eq.~\ref{eq:d} to select the decision for testing data $\bm{x}_i$.

\section{On Computation of Inference Step}
\subsection{Train-test Discrepancy\label{subsec:dis}}
At the inference step, we exploit the importance sampling to convert $p(\mathcal{D}_{test})$ to $p(\mathcal{D}_{train})$ and obtain a discrepancy ratio $p_{test}(\bm{x},y)/p_{train}(\bm{x},y)$. Recall that in long-tailed distribution, the training and testing data are semantically identical, and thus the model prediction must be the same for an input regardless of being in the training or testing set (i.e., $p_{train}(y|\bm{x},\bm{\theta})=p_{test}(y|\bm{x},\bm{\theta})$). Therefore, the discrepancy ratio can be further simplified by:
\begin{equation}
    \frac{p_{test}(\bm{x},y)}{p_{train}(\bm{x},y)}=\frac{p_{test}(y)p_{test}(\bm{x}|y)}{p_{train}(y)p_{train}(\bm{x}|y)}=\frac{p_{test}(y)}{p_{train}(y)},
\end{equation}
which only depends on the class probabilities of training and testing data. Since we assume a uniform distribution for the testing set in long-tailed data, the probability $p_{test}(\bm{y})$ would be a constant for all $\bm{x}$, and thus the discrepancy ratio is equivalent to:
\begin{equation}
    \frac{p_{test}(y)}{p_{train}(y)}\propto\frac{1}{p_{train}(y)}\propto\frac{1}{f(n_y)},
\end{equation}
where $f$ is an increasing function and $n_y$ refers to the number of samples in the class $y$. We introduce the notation of $f(n_y)$ because the class probability only depend on the number of samples in this class.

The choices of $f$ can determine different strategies used by previous re-balancing methods in long-tailed classification. For example, $f(n_y)=n_y^\gamma$ is the most conventional choice with a sensitivity factor $\gamma$ to control the importance of head classes~\cite{huang2016learning,wang2017learning,pan2021model}; $f(n_y)=(1-\beta^{n_y})/(1-\beta)$ is the effective number which considers data overlap~\cite{cui2019class}. A detailed analysis on the choice of discrepancy ratios will be conducted in Section~\ref{subsec:abl}. Notably, our framework is compatible with all previous re-balancing methods as long as they can be expressed in the form of $1/f(n_y)$.

\subsection{Particle-based Variational Distribution}
To pursue the efficiency of model architecture, we use particle optimization~\cite{liu2016stein,d2021repulsive} to obtain the variational distribution: $q(\bm{\theta})=\sum_{j=1}^Mw_j\cdot\delta(\bm{\theta}-\bm{\theta}_j)$, where $\{w_j\}_{j=1}^M$ are normalized weights which hold $\sum_{j=1}^Mw_j=1$, and $\delta(\cdot)$ is the Dirac delta function. The ``particles" $\{\bm{\theta}_j\}_{j=1}^M$ are implemented by ensemble model, which has been empirically explored on the long-tailed data~\cite{wang2020long,li2022trustworthy}. Our formulation gives theoretical justification to ensemble approaches in long-tailed problems: Due to the scarcity of tailed data, there is not enough evidence to support a single solution, leading to many equally good solutions (which give complementary predictions) in the loss landscape. Thus, estimating the full posterior is essential to provide a comprehensive characterization of the solution space. Particle optimization reduces the cost of Bayesian inference and is more efficient than variational inference and Markov chain Monte Carlo (MCMC), especially on high-dimensional and multimodal distributions. Besides, the computational cost of our method can be further reduced by leveraging recent techniques, such as partially being Bayesian in model architectures~\cite{kristiadi2020being}. 

\subsection{Repulsive Regularization}
In Eq.~\ref{eq:obj}, the regularization term $KL(q(\bm{\theta})||p(\bm{\theta}))$ guarantees the variational distribution to approach the posterior as training proceeds. If we assume the prior $p(\bm{\theta})$ to be Gaussian, the regularization can be extended to:
\begin{equation}
\begin{aligned}
    KL&(q(\bm{\theta})||p(\bm{\theta})) \\
    &=\lambda\int_{\Theta}||\bm{\theta}||^2\cdot q(\bm{\theta})d\bm{\theta}+\int_{\Theta}q(\bm{\theta})\log{q(\bm{\theta})}d\bm{\theta} \\
    &=\lambda\cdot\frac{1}{M}\sum_{j=1}^M||\bm{\theta}_j||^2-H(\bm{\theta}),
\end{aligned}
\end{equation}
where $\lambda$ is a constant, $\Theta$ is the parameter space, and $H(\bm{\theta})$ is the entropy of $\bm{\theta}$. The $L_2$-regularization prevents the model from over-fitting and the entropy term applies a \emph{repulsive force} to the particles to promote their diversity, pushing the particles to the target posterior~\cite{d2021repulsive}. A simple approximation for the entropy is used in this paper:
\begin{equation}
    H(\bm{\theta})\propto\frac{1}{2}\log{|\hat{\Sigma}_{\bm{\theta}}|},
\end{equation}
where $\hat{\Sigma}_{\bm{\theta}}$ is the covariance matrix estimated by those particles. Other entropy approximations can also be used. By the technique of SWAG-diagonal covariance~\cite{maddox2019simple}, the covariance matrix can then be directly computed by: $\hat{\Sigma}_{\bm{\theta}}=diag(\overline{\bm{\theta}^2}-\overline{\bm{\theta}}^2)$.

Overall, the regularization term is a combination of $L_2$ weight decay and repulsive force, and is computed by:
\begin{equation}
    KL(q(\bm{\theta})||p(\bm{\theta}))\propto\frac{\lambda}{M}\sum_{j=1}^M||\bm{\theta}_j||^2-\frac{1}{2}\sum_k\log{(\overline{\bm{\theta}^2}-\overline{\bm{\theta}}^2)_k}.
\end{equation}
Our regularization is different from existing diversity regularization~\cite{wang2020long}, and is more principled and naturally derived from the integrated gain. 
 
In summary, our method, with principled design and theoretical justification, is essentially cheap and easy to implement and can be used as a drop-in replacement for existing re-balancing and ensemble methods in general long-tailed problems. We outline our algorithm in~Appendix~\ref{appendix:algo}.

\section{Experiments}
\subsection{Experimental Settings}
\paragraph{Datasets.} We use three long-tailed image datasets. CIFAR-10-LT and CIFAR-100-LT~\cite{cui2019class} are sampled from the original CIFAR dataset~\cite{krizhevsky2009learning}. ImageNet-LT~\cite{liu2019large} is sampled from the the dataset of ILSVRC 2012 competition~\cite{deng2009imagenet}, and contains 115.8K images in 1,000 classes.

\paragraph{Evaluation.} The evaluation protocol consists of standard classification accuracy, a newly designed experiment on the False Head Rate (FHR), and calibration with predictive uncertainty. Besides, we conduct several ablation studies to evaluate different choices of implementation and the effectiveness of components in our method. For all quantitative and visual results, we repeatedly run the experiments five times with random initialization to obtain the averaged results and standard deviations to eliminate random error.

\paragraph{Compared Baselines.} We compare our method (LBD) with cross entropy baseline, re-balancing methods (CB Loss~\cite{cui2019class} and LDAM~\cite{cao2019learning}), and ensemble methods (RIDE~\cite{wang2020long} and TLC~\cite{li2022trustworthy}). The numbers of classifiers in all ensemble models are set to be 3. We also compare the Bayesian predictive uncertainty with MCP baseline~\cite{hendrycks2016baseline} and evidential uncertainty~\cite{csensoy2018evidential}. We use $f(n_y)=n_y$ unless otherwise specified. More implementation details are in Appendix.~\ref{appendix:id}.

\subsection{Standard Classification}
Classification accuracy is the most standard benchmark for long-tailed data, where the overall accuracy and accuracies for three class regions are evaluated. Classes are equally split into three class regions (head, med and tail). For example, there are 33, 33 and 34 classes respectively in the head, med and tail regions of CIFAR-100-LT. The classification results are shown in Table~\ref{tab:classify_all} and Table~\ref{tab:classify}. We apply the one-hot utility in our method to accord with standard accuracy metric. Our method consistently outperforms all other compared methods in terms of overall accuracy. For regional accuracies, our method achieves the best performances on all class regions in most cases. In particular, our method significantly outperforms previous methods on the crucial tailed data, while being comparable or even better on med and head classes. These results demonstrate the effectiveness of taking a Bayesian-decision-theory perspective on the long-tailed problem.
\begin{table}[t]\small\setlength\tabcolsep{4pt}\renewcommand{\arraystretch}{0.9}
\centering
\caption{Quantitative results of overall classification accuracy (in percentage). Our method (LBD) performs the best on all datasets.}
\label{tab:classify_all}
\vskip 0.1in
\begin{tabular}{c|ccc}
\hline
\textsc{Method} & CIFAR-10-LT             & CIFAR-100-LT            & \textsc{ImageNet-LT}    \\ \hline
CE              & 73.65$\pm$0.39          & 38.82$\pm$0.52          & 47.80$\pm$0.15          \\
CB Loss         & 77.62$\pm$0.69          & 42.24$\pm$0.41          & 51.70$\pm$0.25          \\
LDAM            & 80.63$\pm$0.69          & 43.13$\pm$0.67          & 51.04$\pm$0.21          \\
RIDE            & 83.11$\pm$0.52          & 48.99$\pm$0.44          & 54.32$\pm$0.54          \\
TLC             & 79.70$\pm$0.65          & 48.75$\pm$0.16          & 55.03$\pm$0.34          \\
LBD             & \textbf{83.75$\pm$0.17} & \textbf{50.24$\pm$0.70} & \textbf{55.73$\pm$0.17} \\ \hline
\end{tabular}
\end{table}

\begin{table}[t]\small\renewcommand{\arraystretch}{0.9}
\centering
\caption{Quantitative results of classification accuracies on three class regions. LBD outperforms previous methods in all class regions in most cases, especially on tailed data.}
\label{tab:classify}
\vskip 0.15in
\begin{tabular}{c|c|ccc}
\hline
\multirow{2}{*}{$\mathcal{D}$}                & \multirow{2}{*}{\textsc{Method}} & \multicolumn{3}{c}{ACC (\%) $\uparrow$}                                     \\
                                              &                                  & \textsc{head}           & \textsc{med}            & \textsc{tail}           \\ \hline\hline
\multirow{6}{*}{\rotatebox{90}{CIFAR-10-LT}}  & CE                               & 93.22$\pm$0.26          & 74.27$\pm$0.42          & 58.51$\pm$0.62          \\
                                              & CB Loss                          & 91.70$\pm$0.57          & 75.41$\pm$0.76          & 68.73$\pm$1.52          \\
                                              & LDAM                             & 90.03$\pm$0.47          & 75.88$\pm$0.81          & 77.14$\pm$1.61          \\
                                              & RIDE                             & \textbf{91.49$\pm$0.40} & \textbf{79.39$\pm$0.61} & 79.62$\pm$1.56          \\
                                              & TLC                              & 89.47$\pm$0.33          & 74.33$\pm$0.96          & 76.39$\pm$0.98          \\
                                              & LBD                              & 90.49$\pm$0.60          & 78.89$\pm$0.87          & \textbf{82.33$\pm$1.16} \\ \hline
\multirow{6}{*}{\rotatebox{90}{CIFAR-100-LT}} & CE                               & 68.30$\pm$0.61          & 38.39$\pm$0.49          & 10.62$\pm$1.23          \\
                                              & CB Loss                          & 62.53$\pm$0.44          & 44.36$\pm$0.96          & 20.50$\pm$0.51          \\
                                              & LDAM                             & 63.58$\pm$0.93          & 42.90$\pm$1.03          & 23.50$\pm$1.28          \\
                                              & RIDE                             & 69.11$\pm$0.54          & 49.70$\pm$0.59          & 28.78$\pm$1.52          \\
                                              & TLC                              & 69.43$\pm$0.36          & 49.02$\pm$0.94          & 28.40$\pm$0.72          \\
                                              & LBD                              & \textbf{69.92$\pm$0.77} & \textbf{51.07$\pm$0.82} & \textbf{30.34$\pm$1.49} \\ \hline
\multirow{6}{*}{\rotatebox{90}{ImageNet-LT}}  & CE                               & 53.46$\pm$0.36          & 45.92$\pm$0.19          & 44.03$\pm$0.24          \\
                                              & CB Loss                          & 57.62$\pm$0.46          & 49.19$\pm$0.21          & 48.29$\pm$0.41          \\
                                              & LDAM                             & 57.66$\pm$0.40          & 48.26$\pm$0.19          & 47.21$\pm$0.22          \\
                                              & RIDE                             & 60.88$\pm$0.71          & 51.35$\pm$0.44          & 50.74$\pm$0.62          \\
                                              & TLC                              & 61.19$\pm$0.53          & 52.35$\pm$0.31          & 51.56$\pm$0.35          \\
                                              & LBD                              & \textbf{62.18$\pm$0.28} & \textbf{53.06$\pm$0.22} & \textbf{51.98$\pm$0.40} \\ \hline
\end{tabular}
\end{table}

\subsection{Tail-sensitive Classification with False Head Rate\label{subsec:fhr}}
As discussed previously, classifying tailed samples into head classes would often induce negative consequences in real-world applications. Therefore, we are interested in quantifying how likely it will happen, and further evaluating the tail sensitivity of the compared methods. Inspired by the false positive rate, we define the \emph{False Head Rate} (FHR) as:
\begin{equation}
    FHR=\frac{TH}{TT+TH},
\end{equation}
where $TH$ is the number of samples that are labeled as tail but predicted as head, and $TT$ is the number of samples that are labeled and predicted as tail. We also consider different settings of tail region, and select the last 25\%, 50\% and 75\% classes as tail. We apply the tail-sensitive utility in Fig.~\ref{fig:u}(b) to our method. From Table~\ref{tab:fhr}, we observe significant improvements of LBD over previous methods under all settings, especially on the relatively small CIFAR datasets, which means that the ``false head risk" is more severe on smaller datasets with scarce tailed samples. This shows the importance of taking the decision loss into account and also demonstrates the flexibility of our framework which is compatible with different utilities, leading to better performance for different types of tasks. In contrast, previous methods do not coonsider decision loss and may result in undesirable consequences when some type of errors have high costs.  
\begin{table}[t]\scriptsize\setlength\tabcolsep{5pt}
\centering
\caption{Quantitative results of False Head Rates under three different tail ratios and their averaged results. LBD consistently achieves the lowest rates under different scenarios.}
\label{tab:fhr}
\vskip 0.15in
\begin{tabular}{c|c|cccc}
\hline
\multirow{2}{*}{$\mathcal{D}$}                & \multirow{2}{*}{\textsc{Method}} & \multicolumn{4}{c}{FHR (\%) @\textsc{tail ratio} $\downarrow$}                                        \\
                                              &                                  & \textsc{25\%}           & \textsc{50\%}           & \textsc{75\%}           & \textsc{avg}            \\ \hline\hline
\multirow{6}{*}{\rotatebox{90}{CIFAR-10-LT}}  & CE                               & 21.10$\pm$0.43          & 37.87$\pm$0.57          & 48.75$\pm$1.39          & 35.91$\pm$0.54          \\
                                              & CB Loss                          & 14.84$\pm$0.93          & 27.98$\pm$1.44          & 33.93$\pm$1.60          & 25.58$\pm$1.27          \\
                                              & LDAM                             & 10.05$\pm$1.01          & 19.64$\pm$1.66          & 21.37$\pm$2.10          & 17.02$\pm$1.56          \\
                                              & RIDE                             & 8.94$\pm$0.66           & 17.80$\pm$1.39          & 19.77$\pm$3.20          & 15.50$\pm$1.68          \\
                                              & TLC                              & 10.42$\pm$0.64          & 20.27$\pm$0.77          & 22.24$\pm$1.53          & 17.64$\pm$0.93          \\
                                              & LBD                              & \textbf{4.99$\pm$0.32}  & \textbf{11.76$\pm$0.29} & \textbf{11.01$\pm$1.28} & \textbf{9.25$\pm$0.49}  \\ \hline
\multirow{6}{*}{\rotatebox{90}{CIFAR-100-LT}} & CE                               & 45.53$\pm$1.54          & 73.03$\pm$1.59          & 91.30$\pm$1.24          & 69.95$\pm$1.40          \\
                                              & CB Loss                          & 24.88$\pm$0.34          & 48.41$\pm$1.24          & 74.38$\pm$1.47          & 49.22$\pm$0.83          \\
                                              & LDAM                             & 21.22$\pm$0.99          & 43.04$\pm$1.18          & 65.62$\pm$1.31          & 43.29$\pm$1.04          \\
                                              & RIDE                             & 18.83$\pm$0.70          & 39.50$\pm$1.53          & 62.01$\pm$2.70          & 40.11$\pm$1.62          \\
                                              & TLC                              & 21.18$\pm$0.54          & 41.15$\pm$0.55          & 61.34$\pm$1.03          & 41.22$\pm$0.55          \\
                                              & LBD                              & \textbf{15.39$\pm$0.57} & \textbf{31.34$\pm$0.55} & \textbf{49.51$\pm$1.45} & \textbf{32.08$\pm$0.78} \\ \hline
\multirow{6}{*}{\rotatebox{90}{ImageNet-LT}}  & CE                               & 3.99$\pm$0.08           & 12.77$\pm$0.29          & 30.99$\pm$0.40          & 15.92$\pm$0.17          \\
                                              & CB Loss                          & 3.66$\pm$0.17           & 11.80$\pm$0.12          & 29.39$\pm$0.28          & 14.95$\pm$0.15          \\
                                              & LDAM                             & 4.17$\pm$0.19           & 12.73$\pm$0.28          & 29.90$\pm$0.41          & 15.60$\pm$0.21          \\
                                              & RIDE                             & 3.62$\pm$0.18           & 11.42$\pm$0.27          & 26.92$\pm$0.33          & 13.99$\pm$0.24          \\
                                              & TLC                              & 3.47$\pm$0.13           & 11.49$\pm$0.13          & 27.12$\pm$0.13          & 14.03$\pm$0.09          \\
                                              & LBD                              & \textbf{2.70$\pm$0.09}  & \textbf{9.68$\pm$0.25}  & \textbf{24.42$\pm$0.19} & \textbf{12.27$\pm$0.08} \\ \hline
\end{tabular}
\end{table}

\subsection{Calibration}
In our method, the predictive uncertainty can be naturally obtained by the entropy of predictive distribution~\cite{malinin2018predictive}. For the compared uncertainty algorithms, MCP is a trivial baseline which obtains uncertainty scores from the maximum value of softmax distribution, which is added to the RIDE~\cite{wang2020long} backbone; evidential uncertainty is rooted in the subjective logic~\cite{audun2018subjective}, and is introduced to long-tailed classification by TLC~\cite{li2022trustworthy}. We evaluate the uncertainty algorithms with AUC~\cite{mcclish1989analyzing} and ECE~\cite{naeini2015obtaining}, which are shown in Table~\ref{tab:u}. Our Bayesian predictive uncertainty outperforms the other two counterparts and has a remarkable advantage on the ECE metric, demonstrating the superiority of using principled Bayesian uncertainty quantification.  
\begin{table}[t]\small\setlength\tabcolsep{5pt}\renewcommand{\arraystretch}{0.9}
\centering
\caption{Quantitative results of calibration of different uncertainty algorithms. LBD outperforms previous methods remarkably on both metrics and all datasets.}
\label{tab:u}
\vskip 0.15in
\begin{tabular}{c|c|cc}
\hline
\textsc{Dataset}              & \textsc{Algorithm} & AUC (\%) $\uparrow$     & ECE (\%) $\downarrow$   \\ \hline\hline
\multirow{3}{*}{CIFAR-10-LT}  & MCP                & 79.98$\pm$0.10          & 14.33$\pm$0.37          \\
                              & evidential         & 83.20$\pm$0.59          & 13.24$\pm$0.55          \\
                              & Bayesian           & \textbf{86.83$\pm$0.68} & \textbf{9.84$\pm$0.17}  \\ \hline
\multirow{3}{*}{CIFAR-100-LT} & MCP                & 80.48$\pm$0.51          & 23.75$\pm$0.51          \\
                              & evidential         & 77.37$\pm$0.33          & 21.64$\pm$0.47          \\
                              & Bayesian           & \textbf{81.24$\pm$0.25} & \textbf{10.35$\pm$0.28} \\ \hline
\multirow{3}{*}{ImageNet-LT}  & MCP                & 84.02$\pm$0.24          & 18.35$\pm$0.12          \\
                              & evidential         & 81.45$\pm$0.13          & 15.29$\pm$0.12          \\
                              & Bayesian           & \textbf{84.45$\pm$0.09} & \textbf{8.72$\pm$0.13}  \\ \hline
\end{tabular}
\end{table}

\subsection{Ablation Studies\label{subsec:abl}}
\paragraph{Utility Function.} The effectiveness of tail-sensitive utility is shown in Table~\ref{tab:utility}, where we compare the one-hot and tail-sensitive utilities in terms of False Head Rate and classification accuracy. By applying the tail-sensitive utility, the performances on False Head Rate can be significantly improved (18.00\%) with negligible drop on the classification accuracy (0.04\%).
\begin{table*}[t]\renewcommand{\arraystretch}{0.85}
\centering
\caption{Ablation study on the choice of utility function, compared by False Head Rates and classification accuracy on CIFAR-100-LT.}
\label{tab:utility}
\vskip 0.15in
\begin{tabular}{c|ccccc|cc}
\hline
\multirow{2}{*}{\textsc{Utility}} & \multicolumn{4}{c}{FHR (\%) @\textsc{tail ratio} $\downarrow$}                                        & \multirow{2}{*}{\textsc{Better} (\%)} & \multirow{2}{*}{ACC (\%) $\uparrow$} & \multirow{2}{*}{\textsc{Worse} (\%)} \\
                                  & 25\%                    & 50\%                    & 75\%                    & \textsc{avg}            &                                       &                                      &                                      \\ \hline
one-hot                           & 18.55$\pm$0.38          & 38.62$\pm$0.62          & 60.17$\pm$1.48          & 39.12$\pm$0.72          & \multirow{2}{*}{18.00}                & \textbf{49.91$\pm$0.33}              & \multirow{2}{*}{0.04}                \\
tail-sensitive                    & \textbf{15.39$\pm$0.57} & \textbf{31.34$\pm$0.55} & \textbf{49.51$\pm$1.45} & \textbf{32.08$\pm$0.78} &                                       & 49.89$\pm$0.19                       &                                      \\ \hline
\end{tabular}
\end{table*}

\begin{table*}[t]\renewcommand{\arraystretch}{0.85}
\centering
\caption{Ablation study on the choice of discrepancy ratio, compared by classification accuracies on CIFAR-100-LT.}
\label{tab:dis}
\vskip 0.15in
\begin{tabular}{ccc|ccc|c}
\hline
\multicolumn{3}{c|}{\multirow{2}{*}{\textsc{Discrepancy ratio}}}                              & \multicolumn{3}{c|}{\textsc{Weight value}}                        & \multirow{2}{*}{ACC (\%) $\uparrow$} \\
\multicolumn{3}{c|}{}                                                                         & \textsc{first class} & \textsc{last class} & \textsc{growth} (\%) &                                      \\ \hline
\multicolumn{1}{c|}{linear~\cite{wang2017learning}}    & \multirow{5}{*}{$f(n_y)=$} & $n_y$                           & 0.0020               & 0.1667              & \textbf{8250}        & \textbf{50.17$\pm$0.25}              \\
\multicolumn{1}{c|}{effective~\cite{cui2019class}} &                            & $\frac{1-\beta^{n_y}}{1-\beta}$ & 0.0023               & 0.1669              & 7297                 & 49.90$\pm$0.36                       \\
\multicolumn{1}{c|}{sqrt~\cite{pan2021model}}      &                            & $\sqrt{n_y}$                    & 0.0447               & 0.4082              & 814                  & 47.03$\pm$0.30                       \\
\multicolumn{1}{c|}{log}       &                            & $\log{n_y}$                     & 0.1609               & 0.5581              & 247                  & 45.26$\pm$0.51                       \\
\multicolumn{1}{c|}{plain}     &                            & constant                        & 1.0000               & 1.0000              & 0                    & 43.27$\pm$0.30                       \\ \hline
\end{tabular}
\end{table*}

\begin{table}[t]\setlength\tabcolsep{5pt}\renewcommand{\arraystretch}{0.85}
\centering
\caption{Ablation study on repulsive force, compared by uncertainty calibration on CIFAR-100-LT.}
\label{tab:repulsive}
\vskip 0.15in
\begin{tabular}{c|ccc}
\hline
\begin{tabular}[c]{@{}c@{}}\textsc{Repulsive}\\ \textsc{force}\end{tabular} & AUC (\%) $\uparrow$     & ECE (\%) $\downarrow$   & ACC (\%) $\uparrow$     \\ \hline
$\checkmark$                                                                & \textbf{81.24$\pm$0.25} & \textbf{10.35$\pm$0.28} & \textbf{50.24$\pm$0.70} \\
$\times$                                                                    & 75.94$\pm$0.56          & 13.40$\pm$0.80          & 50.15$\pm$0.41          \\ \hline
\end{tabular}
\end{table}

\begin{figure}[ht]
    \centering
    \includegraphics[width=8cm,height=5.5cm]{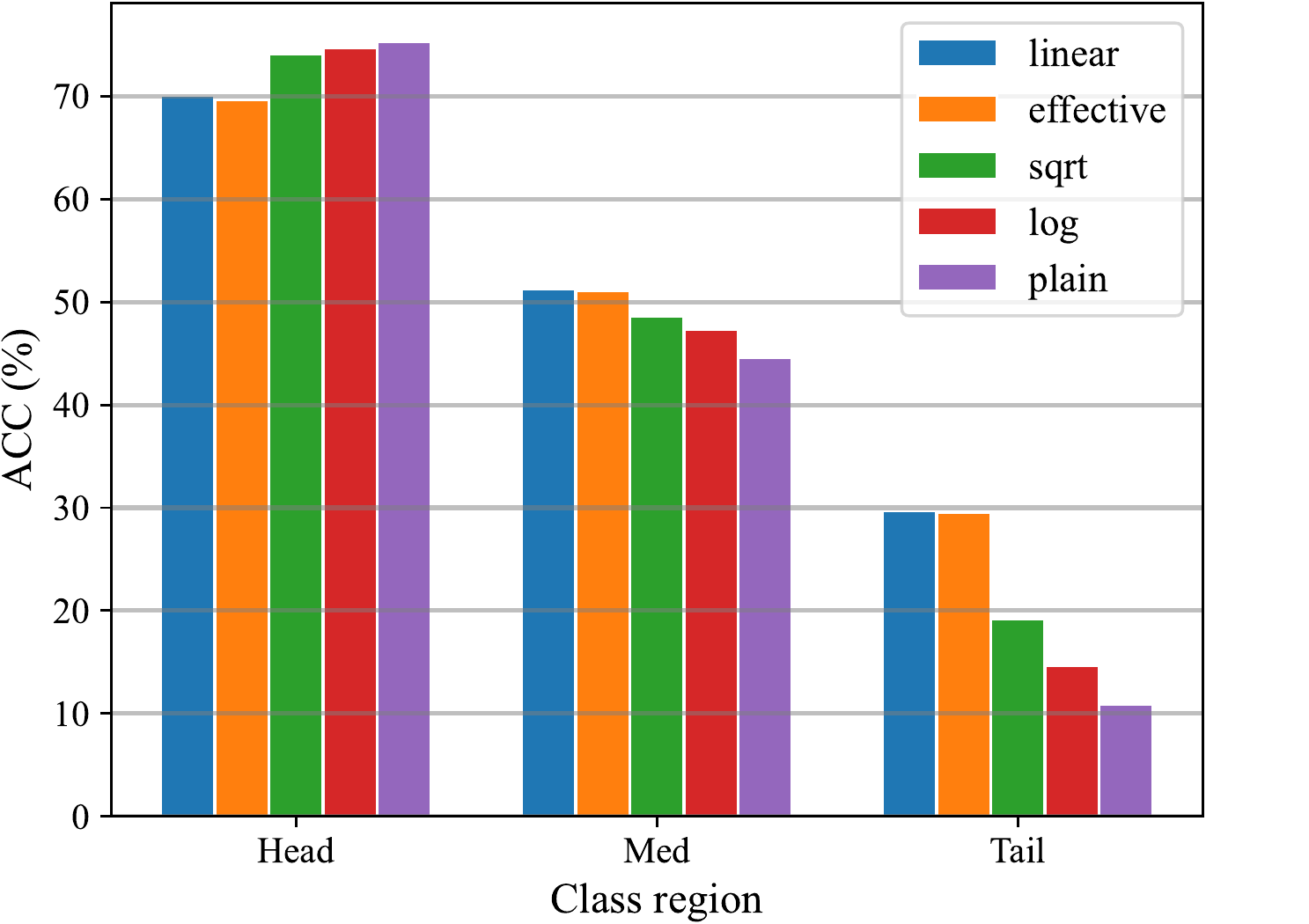}
    \caption{Visual results of classification with respect to the choice of discrepancy ratio on CIFAR-100-LT.}
    \label{fig:dis}
\end{figure}

\paragraph{Train-test Discrepancy.} We compare five different forms of discrepancy ratio in terms of classification accuracy in Table~\ref{tab:dis} and Fig.~\ref{fig:dis}. We also analyze the properties of the compared discrepancy ratios. The differences of $f(n_y)$ show up when $n_y$ is large, and it can be measured by the growth rate of weight values (i.e., $1/f(n_y)$) between the first and the last class. We find that as the growth rate becomes larger, the overall accuracy will be better accordingly, which shows the severity of class imbalance.

For the classification accuracies of three class regions, Fig.~\ref{fig:dis} shows similar results on the relationship between growth rate and the tail accuracy. As the growth rate becomes larger, the tail and med ACC will both become significantly better despite the slight drop on head ACC, which is consistent with the overall improvement. Based on these results, we suggest using $f(n_y) = n_y$ in general.

\paragraph{Repulsive Force.} We evaluate the effectiveness of repulsive force in Table~\ref{tab:repulsive}. The repulsive force effectively pushes the particles to the target posterior and avoids collapsing into the same solution. Therefore, with the repulsive force, better predictive distributions can be learned, and thus better predictive uncertainty can be obtained. Besides, the repulsive force can also improve the accuracy by promoting the diversity of particles.

\begin{figure}[ht]
    \centering
    \includegraphics[width=8cm,height=5.5cm]{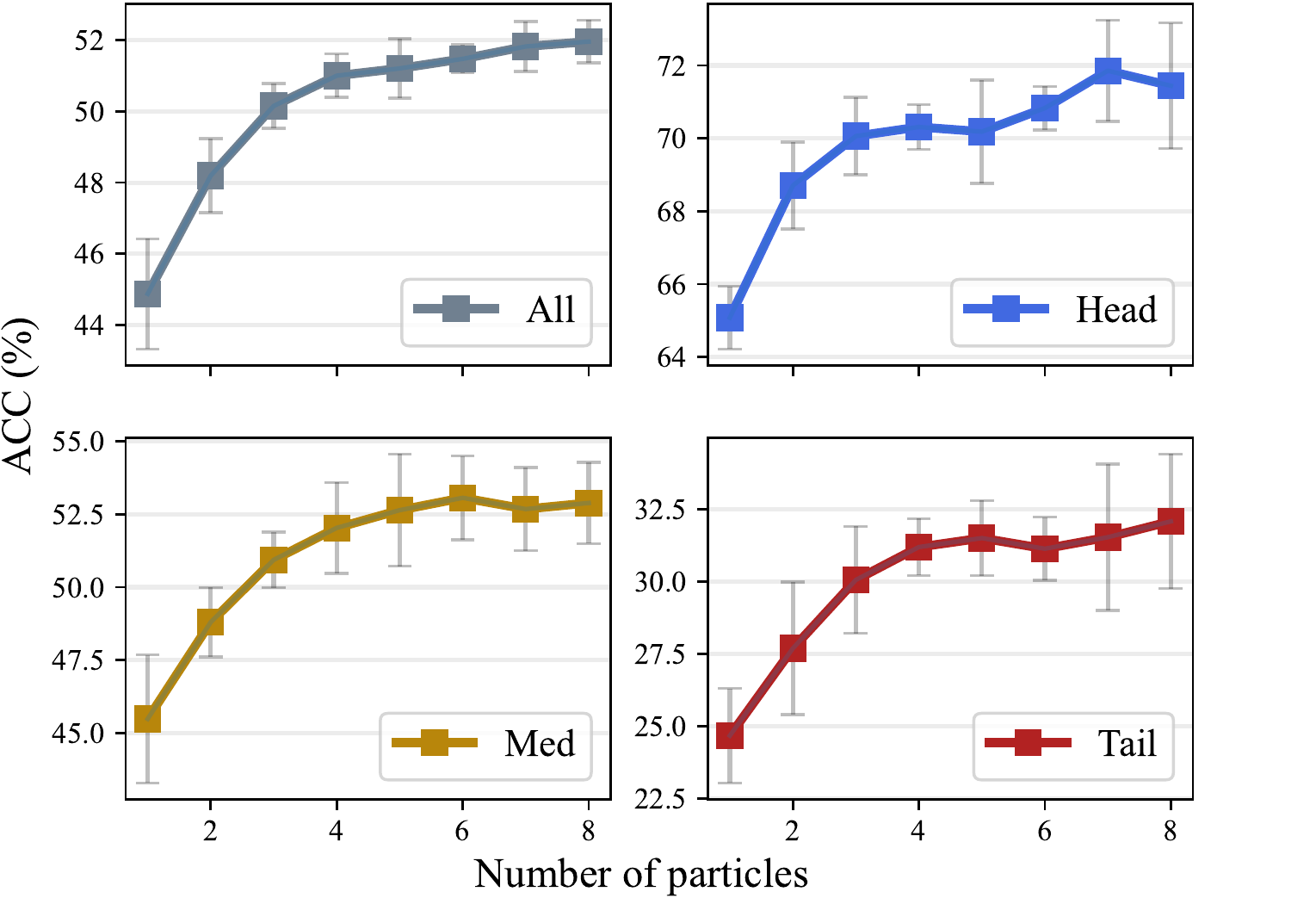}
    \caption{Visual results of classification with respect to the number of particles on CIFAR-100-LT.}
    \label{fig:particle}
\end{figure}

\paragraph{Number of Particles.} Generally, using more classifiers in ensembles will induce better performances. However, we also need to balance the performance with the computational cost. We visualize the classification accuracies under different numbers of particles in Fig.~\ref{fig:particle}. The error bars are scaled to be $2\sigma$, where $\sigma$ is the standard deviation from repeated experiments. The accuracy curves are all logarithm-like and the accuracy improvement is hardly noticeable for more than six particles. However, the computational cost is increasing in a linear speed. Therefore, we recommend using no more than six particles in practice for a desirable performance-cost trade-off.

\section{Conclusion and Future Directions}
In this paper, we propose Long-tailed Bayesian Decision (LBD), a principled framework for solving long-tailed problems, with both theoretical explanation and strong empirical performance. Based on Bayesian Decision Theory, LBD unifies data distribution, posterior inference, and decision-making and further provides theoretical justification for existing techniques such as re-balancing and ensemble. In LBD, we introduce the integrated risk as the objective, find a tractable variational lower bound to optimize this objective, and apply particle optimization to efficiently estimate the complex posterior. For the real-world scenario with tail sensitivity risk, we design a tail-sensitive utility to pursue a better False Head Rate. In experiments, we evaluate our framework on standard classification, tail-sensitive classification, calibration, and ablation studies. Our framework outperforms the current SOTA even on large-scale real-world datasets like ImageNet. 

Our method is simple to use in general long-tailed problems, providing superior accuracy on all types of classes and uncertainty estimation. We believe there is considerable space for future developments that build upon our method and we list a few below:
  
\paragraph{Long-tailed Regression.} Long-tail problem also exists in regression, where the distribution of targets can be heavily imbalanced. With some adjustments on the decision gain, our framework might also be adapted to regression.

\paragraph{Utility Function.} Beyond long-tailed classification, there are other tasks which also need specific utility functions. For example, we might have to separately deal with the relationship between categories due to their semantic connections. In this case, all of the values in the utility matrix will need re-calculating.

\paragraph{Dataset Shift.} In more general dataset shift scenarios like out-of-distribution data, the assumption about semantically identical training and testing sets will be no longer valid. Another example is about the distribution of testing data. If it is assumed to be not uniform, the discrepancy ratio $p_{test}(y)/p_{train}(y)$ will no longer be expressed in the form $1/f(n_y)$, but a more general form.

\bibliography{main}

\begin{thebibliography}{65}
\providecommand{\natexlab}[1]{#1}
\providecommand{\url}[1]{\texttt{#1}}
\expandafter\ifx\csname urlstyle\endcsname\relax
  \providecommand{\doi}[1]{doi: #1}\else
  \providecommand{\doi}{doi: \begingroup \urlstyle{rm}\Url}\fi

\bibitem[Audun(2018)]{audun2018subjective}
Audun, J.
\newblock \emph{Subjective Logic: A formalism for reasoning under uncertainty}.
\newblock Springer, 2018.

\bibitem[Berger(1985)]{berger2013statistical}
Berger, J.~O.
\newblock \emph{Statistical decision theory and Bayesian analysis}.
\newblock Springer Science \& Business Media, 1985.

\bibitem[Bottou(1998)]{bottou1998online}
Bottou, L.
\newblock Online algorithms and stochastic approxima-p tions.
\newblock \emph{Online learning and neural networks}, 1998.

\bibitem[Cao et~al.(2019)Cao, Wei, Gaidon, Arechiga, and Ma]{cao2019learning}
Cao, K., Wei, C., Gaidon, A., Arechiga, N., and Ma, T.
\newblock Learning imbalanced datasets with label-distribution-aware margin
  loss.
\newblock \emph{Advances in neural information processing systems}, 32, 2019.

\bibitem[Chen et~al.(2018)Chen, Zhang, Wang, Li, and Chen]{chen2018unified}
Chen, C., Zhang, R., Wang, W., Li, B., and Chen, L.
\newblock A unified particle-optimization framework for scalable bayesian
  sampling.
\newblock \emph{stat}, 1050:\penalty0 10, 2018.

\bibitem[Chu et~al.(2020)Chu, Bian, Liu, and Ling]{chu2020feature}
Chu, P., Bian, X., Liu, S., and Ling, H.
\newblock Feature space augmentation for long-tailed data.
\newblock In \emph{Computer Vision--ECCV 2020: 16th European Conference,
  Glasgow, UK, August 23--28, 2020, Proceedings, Part XXIX 16}, pp.\  694--710.
  Springer, 2020.

\bibitem[Cobb et~al.(2018)Cobb, Roberts, and Gal]{cobb2018loss}
Cobb, A.~D., Roberts, S.~J., and Gal, Y.
\newblock Loss-calibrated approximate inference in bayesian neural networks.
\newblock \emph{arXiv preprint arXiv:1805.03901}, 2018.

\bibitem[Cui et~al.(2019)Cui, Jia, Lin, Song, and Belongie]{cui2019class}
Cui, Y., Jia, M., Lin, T.-Y., Song, Y., and Belongie, S.
\newblock Class-balanced loss based on effective number of samples.
\newblock In \emph{Proceedings of the IEEE/CVF conference on computer vision
  and pattern recognition}, pp.\  9268--9277, 2019.

\bibitem[D'Angelo \& Fortuin(2020)D'Angelo and Fortuin]{d2020annealed}
D'Angelo, F. and Fortuin, V.
\newblock Annealed stein variational gradient descent.
\newblock In \emph{Third Symposium on Advances in Approximate Bayesian
  Inference}, 2020.

\bibitem[D'Angelo \& Fortuin(2021)D'Angelo and Fortuin]{d2021repulsive}
D'Angelo, F. and Fortuin, V.
\newblock Repulsive deep ensembles are bayesian.
\newblock \emph{Advances in Neural Information Processing Systems},
  34:\penalty0 3451--3465, 2021.

\bibitem[Dempster et~al.(1977)Dempster, Laird, and Rubin]{dempster1977maximum}
Dempster, A.~P., Laird, N.~M., and Rubin, D.~B.
\newblock Maximum likelihood from incomplete data via the em algorithm.
\newblock \emph{Journal of the Royal Statistical Society: Series B
  (Methodological)}, 39\penalty0 (1):\penalty0 1--22, 1977.

\bibitem[Deng et~al.(2009)Deng, Dong, Socher, Li, Li, and
  Fei-Fei]{deng2009imagenet}
Deng, J., Dong, W., Socher, R., Li, L.-J., Li, K., and Fei-Fei, L.
\newblock Imagenet: A large-scale hierarchical image database.
\newblock In \emph{2009 IEEE conference on computer vision and pattern
  recognition}, pp.\  248--255. Ieee, 2009.

\bibitem[Dietterich(2000)]{dietterich2000ensemble}
Dietterich, T.~G.
\newblock Ensemble methods in machine learning.
\newblock In \emph{International workshop on multiple classifier systems}, pp.\
   1--15. Springer, 2000.

\bibitem[Ganaie et~al.(2021)Ganaie, Hu, et~al.]{ganaie2021ensemble}
Ganaie, M.~A., Hu, M., et~al.
\newblock Ensemble deep learning: A review.
\newblock \emph{arXiv preprint arXiv:2104.02395}, 2021.

\bibitem[Geng et~al.(2020)Geng, Huang, and Chen]{geng2020recent}
Geng, C., Huang, S.-j., and Chen, S.
\newblock Recent advances in open set recognition: A survey.
\newblock \emph{IEEE transactions on pattern analysis and machine
  intelligence}, 43\penalty0 (10):\penalty0 3614--3631, 2020.

\bibitem[Han et~al.(2005)Han, Wang, and Mao]{han2005borderline}
Han, H., Wang, W.-Y., and Mao, B.-H.
\newblock Borderline-smote: a new over-sampling method in imbalanced data sets
  learning.
\newblock In \emph{International conference on intelligent computing}, pp.\
  878--887. Springer, 2005.

\bibitem[Hand(2006)]{hand2006classifier}
Hand, D.~J.
\newblock Classifier technology and the illusion of progress.
\newblock \emph{Statistical science}, 21\penalty0 (1):\penalty0 1--14, 2006.

\bibitem[Hendrycks \& Gimpel(2017)Hendrycks and Gimpel]{hendrycks2016baseline}
Hendrycks, D. and Gimpel, K.
\newblock A baseline for detecting misclassified and out-of-distribution
  examples in neural networks.
\newblock \emph{Proceedings of International Conference on Learning
  Representations}, 2017.

\bibitem[Huang et~al.(2016)Huang, Li, Loy, and Tang]{huang2016learning}
Huang, C., Li, Y., Loy, C.~C., and Tang, X.
\newblock Learning deep representation for imbalanced classification.
\newblock In \emph{Proceedings of the IEEE conference on computer vision and
  pattern recognition}, pp.\  5375--5384, 2016.

\bibitem[Jensen(1906)]{jensen1906fonctions}
Jensen, J. L. W.~V.
\newblock Sur les fonctions convexes et les in{\'e}galit{\'e}s entre les
  valeurs moyennes.
\newblock \emph{Acta mathematica}, 30\penalty0 (1):\penalty0 175--193, 1906.

\bibitem[Kim et~al.(2020)Kim, Jeong, and Shin]{kim2020m2m}
Kim, J., Jeong, J., and Shin, J.
\newblock M2m: Imbalanced classification via major-to-minor translation.
\newblock In \emph{Proceedings of the IEEE/CVF Conference on Computer Vision
  and Pattern Recognition}, pp.\  13896--13905, 2020.

\bibitem[Kloek \& Van~Dijk(1978)Kloek and Van~Dijk]{kloek1978bayesian}
Kloek, T. and Van~Dijk, H.~K.
\newblock Bayesian estimates of equation system parameters: an application of
  integration by monte carlo.
\newblock \emph{Econometrica: Journal of the Econometric Society}, pp.\  1--19,
  1978.

\bibitem[Korba et~al.(2020)Korba, Salim, Arbel, Luise, and
  Gretton]{korba2020non}
Korba, A., Salim, A., Arbel, M., Luise, G., and Gretton, A.
\newblock A non-asymptotic analysis for stein variational gradient descent.
\newblock \emph{Advances in Neural Information Processing Systems},
  33:\penalty0 4672--4682, 2020.

\bibitem[Krishna et~al.(2017)Krishna, Zhu, Groth, Johnson, Hata, Kravitz, Chen,
  Kalantidis, Li, Shamma, et~al.]{krishna2017visual}
Krishna, R., Zhu, Y., Groth, O., Johnson, J., Hata, K., Kravitz, J., Chen, S.,
  Kalantidis, Y., Li, L.-J., Shamma, D.~A., et~al.
\newblock Visual genome: Connecting language and vision using crowdsourced
  dense image annotations.
\newblock \emph{International journal of computer vision}, 123\penalty0
  (1):\penalty0 32--73, 2017.

\bibitem[Kristiadi et~al.(2020)Kristiadi, Hein, and Hennig]{kristiadi2020being}
Kristiadi, A., Hein, M., and Hennig, P.
\newblock Being bayesian, even just a bit, fixes overconfidence in relu
  networks.
\newblock In \emph{International conference on machine learning}, pp.\
  5436--5446. PMLR, 2020.

\bibitem[Krizhevsky \& Hinton(2009)Krizhevsky and
  Hinton]{krizhevsky2009learning}
Krizhevsky, A. and Hinton, G.
\newblock Learning multiple layers of features from tiny images.
\newblock Technical Report~0, University of Toronto, Toronto, Ontario, 2009.

\bibitem[Lacoste-Julien et~al.(2011)Lacoste-Julien, Husz{\'a}r, and
  Ghahramani]{lacoste2011approximate}
Lacoste-Julien, S., Husz{\'a}r, F., and Ghahramani, Z.
\newblock Approximate inference for the loss-calibrated bayesian.
\newblock In \emph{Proceedings of the Fourteenth International Conference on
  Artificial Intelligence and Statistics}, pp.\  416--424. JMLR Workshop and
  Conference Proceedings, 2011.

\bibitem[Lakshminarayanan et~al.(2017)Lakshminarayanan, Pritzel, and
  Blundell]{lakshminarayanan2017simple}
Lakshminarayanan, B., Pritzel, A., and Blundell, C.
\newblock Simple and scalable predictive uncertainty estimation using deep
  ensembles.
\newblock \emph{Advances in Neural Information Processing Systems}, 30, 2017.

\bibitem[Li et~al.(2022)Li, Han, Li, Fu, and Zhang]{li2022trustworthy}
Li, B., Han, Z., Li, H., Fu, H., and Zhang, C.
\newblock Trustworthy long-tailed classification.
\newblock In \emph{Proceedings of the IEEE/CVF Conference on Computer Vision
  and Pattern Recognition}, pp.\  6970--6979, 2022.

\bibitem[Lin et~al.(2014)Lin, Maire, Belongie, Hays, Perona, Ramanan,
  Doll{\'a}r, and Zitnick]{lin2014microsoft}
Lin, T.-Y., Maire, M., Belongie, S., Hays, J., Perona, P., Ramanan, D.,
  Doll{\'a}r, P., and Zitnick, C.~L.
\newblock Microsoft coco: Common objects in context.
\newblock In \emph{European conference on computer vision}, pp.\  740--755.
  Springer, 2014.

\bibitem[Lin et~al.(2017)Lin, Goyal, Girshick, He, and
  Doll{\'a}r]{lin2017focal}
Lin, T.-Y., Goyal, P., Girshick, R., He, K., and Doll{\'a}r, P.
\newblock Focal loss for dense object detection.
\newblock In \emph{Proceedings of the IEEE international conference on computer
  vision}, pp.\  2980--2988, 2017.

\bibitem[Liu et~al.(2018)Liu, Zhuo, Cheng, Zhang, Zhu, and
  Carin]{liu2018accelerated}
Liu, C., Zhuo, J., Cheng, P., Zhang, R., Zhu, J., and Carin, L.
\newblock Accelerated first-order methods on the wasserstein space for bayesian
  inference.
\newblock \emph{stat}, 1050:\penalty0 4, 2018.

\bibitem[Liu et~al.(2019{\natexlab{a}})Liu, Zhuo, Cheng, Zhang, and
  Zhu]{liu2019understanding}
Liu, C., Zhuo, J., Cheng, P., Zhang, R., and Zhu, J.
\newblock Understanding and accelerating particle-based variational inference.
\newblock In \emph{International Conference on Machine Learning}, pp.\
  4082--4092. PMLR, 2019{\natexlab{a}}.

\bibitem[Liu et~al.(2020)Liu, Sun, Han, Dou, and Li]{liu2020deep}
Liu, J., Sun, Y., Han, C., Dou, Z., and Li, W.
\newblock Deep representation learning on long-tailed data: A learnable
  embedding augmentation perspective.
\newblock In \emph{Proceedings of the IEEE/CVF Conference on Computer Vision
  and Pattern Recognition}, pp.\  2970--2979, 2020.

\bibitem[Liu(2017)]{liu2017stein}
Liu, Q.
\newblock Stein variational gradient descent as gradient flow.
\newblock \emph{Advances in neural information processing systems}, 30, 2017.

\bibitem[Liu \& Wang(2016)Liu and Wang]{liu2016stein}
Liu, Q. and Wang, D.
\newblock Stein variational gradient descent: A general purpose bayesian
  inference algorithm.
\newblock \emph{Advances in neural information processing systems}, 29, 2016.

\bibitem[Liu et~al.(2008)Liu, Wu, and Zhou]{liu2008exploratory}
Liu, X.-Y., Wu, J., and Zhou, Z.-H.
\newblock Exploratory undersampling for class-imbalance learning.
\newblock \emph{IEEE Transactions on Systems, Man, and Cybernetics, Part B
  (Cybernetics)}, 39\penalty0 (2):\penalty0 539--550, 2008.

\bibitem[Liu et~al.(2019{\natexlab{b}})Liu, Miao, Zhan, Wang, Gong, and
  Yu]{liu2019large}
Liu, Z., Miao, Z., Zhan, X., Wang, J., Gong, B., and Yu, S.~X.
\newblock Large-scale long-tailed recognition in an open world.
\newblock In \emph{Proceedings of the IEEE/CVF Conference on Computer Vision
  and Pattern Recognition}, pp.\  2537--2546, 2019{\natexlab{b}}.

\bibitem[Maddox et~al.(2019)Maddox, Izmailov, Garipov, Vetrov, and
  Wilson]{maddox2019simple}
Maddox, W.~J., Izmailov, P., Garipov, T., Vetrov, D.~P., and Wilson, A.~G.
\newblock A simple baseline for bayesian uncertainty in deep learning.
\newblock \emph{Advances in Neural Information Processing Systems}, 32, 2019.

\bibitem[Mahajan et~al.(2018)Mahajan, Girshick, Ramanathan, He, Paluri, Li,
  Bharambe, and Van Der~Maaten]{mahajan2018exploring}
Mahajan, D., Girshick, R., Ramanathan, V., He, K., Paluri, M., Li, Y.,
  Bharambe, A., and Van Der~Maaten, L.
\newblock Exploring the limits of weakly supervised pretraining.
\newblock In \emph{Proceedings of the European conference on computer vision
  (ECCV)}, pp.\  181--196, 2018.

\bibitem[Malinin \& Gales(2018)Malinin and Gales]{malinin2018predictive}
Malinin, A. and Gales, M.
\newblock Predictive uncertainty estimation via prior networks.
\newblock \emph{Advances in neural information processing systems}, 31, 2018.

\bibitem[McClish(1989)]{mcclish1989analyzing}
McClish, D.~K.
\newblock Analyzing a portion of the roc curve.
\newblock \emph{Medical decision making}, 9\penalty0 (3):\penalty0 190--195,
  1989.

\bibitem[Menon et~al.(2020)Menon, Jayasumana, Rawat, Jain, Veit, and
  Kumar]{menon2020long}
Menon, A.~K., Jayasumana, S., Rawat, A.~S., Jain, H., Veit, A., and Kumar, S.
\newblock Long-tail learning via logit adjustment.
\newblock In \emph{International Conference on Learning Representations}, 2020.

\bibitem[Morais \& Pillow(2022)Morais and Pillow]{morais2022loss}
Morais, M.~J. and Pillow, J.~W.
\newblock Loss-calibrated expectation propagation for approximate bayesian
  decision-making.
\newblock \emph{arXiv preprint arXiv:2201.03128}, 2022.

\bibitem[Moreno-Torres et~al.(2012)Moreno-Torres, Raeder, Alaiz-Rodr{\'\i}guez,
  Chawla, and Herrera]{moreno2012unifying}
Moreno-Torres, J.~G., Raeder, T., Alaiz-Rodr{\'\i}guez, R., Chawla, N.~V., and
  Herrera, F.
\newblock A unifying view on dataset shift in classification.
\newblock \emph{Pattern recognition}, 45\penalty0 (1):\penalty0 521--530, 2012.

\bibitem[Naeini et~al.(2015)Naeini, Cooper, and
  Hauskrecht]{naeini2015obtaining}
Naeini, M.~P., Cooper, G., and Hauskrecht, M.
\newblock Obtaining well calibrated probabilities using bayesian binning.
\newblock In \emph{Twenty-Ninth AAAI Conference on Artificial Intelligence},
  2015.

\bibitem[Nam et~al.(2023)Nam, Jang, and Lee]{namdecoupled}
Nam, G., Jang, S., and Lee, J.
\newblock Decoupled training for long-tailed classification with stochastic
  representations.
\newblock In \emph{The Eleventh International Conference on Learning
  Representations}, 2023.

\bibitem[Pan et~al.(2021)Pan, Zhang, Li, Hu, Xuan, Changpinyo, Gong, and
  Chao]{pan2021model}
Pan, T.-Y., Zhang, C., Li, Y., Hu, H., Xuan, D., Changpinyo, S., Gong, B., and
  Chao, W.-L.
\newblock On model calibration for long-tailed object detection and instance
  segmentation.
\newblock \emph{Advances in Neural Information Processing Systems},
  34:\penalty0 2529--2542, 2021.

\bibitem[Quinonero-Candela et~al.(2008)Quinonero-Candela, Sugiyama,
  Schwaighofer, and Lawrence]{quinonero2008dataset}
Quinonero-Candela, J., Sugiyama, M., Schwaighofer, A., and Lawrence, N.~D.
\newblock \emph{Dataset shift in machine learning}.
\newblock Mit Press, 2008.

\bibitem[Rahman et~al.(2021)Rahman, Hassan, and Ahad]{rahman2021nurse}
Rahman, A., Hassan, I., and Ahad, M. A.~R.
\newblock Nurse care activity recognition: A cost-sensitive ensemble approach
  to handle imbalanced class problem in the wild.
\newblock In \emph{Adjunct Proceedings of the 2021 ACM International Joint
  Conference on Pervasive and Ubiquitous Computing and Proceedings of the 2021
  ACM International Symposium on Wearable Computers}, pp.\  440--445, 2021.

\bibitem[Reed(2001)]{reed2001pareto}
Reed, W.~J.
\newblock The pareto, zipf and other power laws.
\newblock \emph{Economics letters}, 74\penalty0 (1):\penalty0 15--19, 2001.

\bibitem[Robert et~al.(2007)]{robert2007bayesian}
Robert, C.~P. et~al.
\newblock \emph{The Bayesian choice: from decision-theoretic foundations to
  computational implementation}, volume~2.
\newblock Springer, 2007.

\bibitem[Schervish(2012)]{schervish2012theory}
Schervish, M.~J.
\newblock \emph{Theory of statistics}.
\newblock Springer Science \& Business Media, 2012.

\bibitem[Sengupta et~al.(2016)Sengupta, Chen, Castillo, Patel, Chellappa, and
  Jacobs]{sengupta2016frontal}
Sengupta, S., Chen, J.-C., Castillo, C., Patel, V.~M., Chellappa, R., and
  Jacobs, D.~W.
\newblock Frontal to profile face verification in the wild.
\newblock In \emph{2016 IEEE winter conference on applications of computer
  vision (WACV)}, pp.\  1--9. IEEE, 2016.

\bibitem[{\c{S}}ensoy et~al.(2018){\c{S}}ensoy, Kaplan, and
  Kandemir]{csensoy2018evidential}
{\c{S}}ensoy, M., Kaplan, L., and Kandemir, M.
\newblock Evidential deep learning to quantify classification uncertainty.
\newblock \emph{Advances in Neural Information Processing Systems}, 2018.

\bibitem[Vadera et~al.(2021)Vadera, Ghosh, Ng, and Marlin]{vadera2021post}
Vadera, M.~P., Ghosh, S., Ng, K., and Marlin, B.~M.
\newblock Post-hoc loss-calibration for bayesian neural networks.
\newblock In \emph{Uncertainty in Artificial Intelligence}, pp.\  1403--1412.
  PMLR, 2021.

\bibitem[Van~Horn \& Perona(2017)Van~Horn and Perona]{van2017devil}
Van~Horn, G. and Perona, P.
\newblock The devil is in the tails: Fine-grained classification in the wild.
\newblock \emph{arXiv preprint arXiv:1709.01450}, 2017.

\bibitem[Wang et~al.(2022)Wang, Zhang, Zhu, Zheng, Li, Smola, and
  Wang]{wang2022partial}
Wang, H., Zhang, A., Zhu, Y., Zheng, S., Li, M., Smola, A.~J., and Wang, Z.
\newblock Partial and asymmetric contrastive learning for out-of-distribution
  detection in long-tailed recognition.
\newblock In \emph{International Conference on Machine Learning}, pp.\
  23446--23458. PMLR, 2022.

\bibitem[Wang et~al.(2020)Wang, Lian, Miao, Liu, and Yu]{wang2020long}
Wang, X., Lian, L., Miao, Z., Liu, Z., and Yu, S.
\newblock Long-tailed recognition by routing diverse distribution-aware
  experts.
\newblock In \emph{International Conference on Learning Representations}, 2020.

\bibitem[Wang et~al.(2017)Wang, Ramanan, and Hebert]{wang2017learning}
Wang, Y.-X., Ramanan, D., and Hebert, M.
\newblock Learning to model the tail.
\newblock \emph{Advances in neural information processing systems}, 30, 2017.

\bibitem[Wang et~al.(2018)Wang, Ren, Zhu, and Zhang]{wang2018function}
Wang, Z., Ren, T., Zhu, J., and Zhang, B.
\newblock Function space particle optimization for bayesian neural networks.
\newblock In \emph{International Conference on Learning Representations}, 2018.

\bibitem[Wu et~al.(2020)Wu, Huang, Liu, Wang, and Lin]{wu2020distribution}
Wu, T., Huang, Q., Liu, Z., Wang, Y., and Lin, D.
\newblock Distribution-balanced loss for multi-label classification in
  long-tailed datasets.
\newblock In \emph{European Conference on Computer Vision}, pp.\  162--178.
  Springer, 2020.

\bibitem[Xiang et~al.(2020)Xiang, Ding, and Han]{xiang2020learning}
Xiang, L., Ding, G., and Han, J.
\newblock Learning from multiple experts: Self-paced knowledge distillation for
  long-tailed classification.
\newblock In \emph{European Conference on Computer Vision}, pp.\  247--263.
  Springer, 2020.

\bibitem[Yang et~al.(2022)Yang, Pan, Yang, Shi, Zhou, Zhang, and
  Bian]{yang2022proco}
Yang, Z., Pan, J., Yang, Y., Shi, X., Zhou, H.-Y., Zhang, Z., and Bian, C.
\newblock Proco: Prototype-aware contrastive learning for long-tailed medical
  image classification.
\newblock In \emph{International Conference on Medical Image Computing and
  Computer-Assisted Intervention}, pp.\  173--182. Springer, 2022.

\bibitem[Zhang et~al.(2022)Zhang, Hooi, Lanqing, and Feng]{zhangself}
Zhang, Y., Hooi, B., Lanqing, H., and Feng, J.
\newblock Self-supervised aggregation of diverse experts for test-agnostic
  long-tailed recognition.
\newblock In \emph{Advances in Neural Information Processing Systems}, 2022.

\end{thebibliography}
\bibliographystyle{icml2023}

\newpage
\appendix
\onecolumn
\section{Decision Gain $g(\bm{\theta},d(\bm{x}))$\label{sec:decision_gain}}
As mentioned in the paper, we design the decision gain to be the following:
\begin{equation}
    g(\bm{\theta},d(\bm{x}))\propto -l(\bm{\theta},d(\bm{x})):=\prod_{y'}p(y'|\bm{x},\bm{\theta})^{u(y',d(\bm{x}))}.\label{A1}
\end{equation}
Our design is different from previous work~\cite{cobb2018loss}, which uses
\begin{equation}
    g(\bm{\theta},d(\bm{x})):=\sum_{y'}p(y'|\bm{x},\bm{\theta})u(y',d(\bm{x})).\label{A2}
\end{equation}
Both definitions achieve the goal of averaging the utility over the label distribution $p(y|\bm{x},\bm{\theta})$. However, our design has two advantages: \romannumeral1) Eq.~\ref{A1} is more stable for training. After taking the $\log$, Eq.~\ref{A1} becomes $\sum_{y'}u(y',d(\bm{x}))\log p(y'|\bm{x},\bm{\theta})$ which is a weighted average of the logarithm of probabilities, while Eq.~\ref{A2} becomes $\log\sum_{y'}u(y',d(\bm{x})) p(y'|\bm{x},\bm{\theta})$, which is a weighted average of the probabilities. \romannumeral2) Eq.~\ref{A1} allows for more general and flexible utility functions whereas Eq.~\ref{A2} requires utility $u$ to be positive (otherwise we may not be able to compute the logarithm of Eq.~\ref{A2}). Due to these reasons, we use Eq.~\ref{A1} in this paper.

\section{Algorithm Summary\label{appendix:algo}}
\begin{algorithm}[h]
   \caption{Long-tailed Bayesian Decision}
\begin{algorithmic}
   \STATE {\bfseries Input:} dataset $\mathcal{D}_{train}=\{(\bm{x}_i,y_i)\}_{i=1}^N$, model $\{\bm{\theta}_j\}_{j=1}^M$ ($M$ particles), utility matrix $\bm{U}$
   \STATE Randomly initialize $\{\bm{\theta}_j\}_{j=1}^M$.
   \FOR{$\text{epoch}=1$ {\bfseries to} $T$}
   \STATE Get a minibatch of data from $\mathcal{D}_{train}$.
   \STATE Compute the objective $L(q)$ as in Eq.~\eqref{eq:final_obj} using the minibatch.
   \STATE Update $\{\bm{\theta}_j\}_{j=1}^M$ using stochastic gradient descent.
   \ENDFOR
   \vspace{1em}
   \STATE \emph{// Testing}
   \STATE Obtain the decision $d$ for testing data $x$ as in Eq.~\eqref{eq:d}.
\end{algorithmic}
\end{algorithm}

\section{Inference Step\label{appendix:e}}
\emph{Proof.} We denote the training and testing sets as $\mathcal{D}_{train}=\{\mathcal{X}_{train},\mathcal{Y}_{train}\}$ and $\mathcal{D}_{test}=\{\mathcal{X}_{test},\mathcal{Y}_{test}\}$ respectively. The maximization objective would be:
\begin{equation}
\begin{aligned}
    \log{G(d)}&=\log{\mathbb{E}_{(\bm{x},y)\sim p_{test}(\bm{x},y)}\mathbb{E}_{\bm{\theta}\sim p(\bm{\theta|\mathcal{D}_{test}})}g(\bm{\theta},d(\bm{x}))} \\
    &\overset{\text{(a)}}{\geq}\mathbb{E}_{(\bm{x},y)\sim p_{test}(\bm{x},y)}\log\mathbb{E}_{\bm{\theta}\sim p(\bm{\theta|\mathcal{D}_{test}})}g(\bm{\theta},d(\bm{x})) \\
    &=\mathbb{E}_{(\bm{x},y)\sim p_{test}(\bm{x},y)}\log{\int_{\Theta}q(\bm{\theta})g(\bm{\theta},d(\bm{x}))\frac{p(\bm{\theta}|\mathcal{D}_{test})}{q(\bm{\theta})}d\bm{\theta}} \\
    &\overset{\text{(b)}}{\geq}\mathbb{E}_{(\bm{x},y)\sim p_{test}(\bm{x},y)}\int_{\Theta}q(\bm{\theta})\log{\left[g(\bm{\theta},d(\bm{x}))\frac{p(\bm{\theta}|\mathcal{D}_{test})}{q(\bm{\theta})}\right]}d\bm{\theta}.\label{eq:a_e}
\end{aligned}
\end{equation}
Here, (a) and (b) are by Jensen’s inequality~\cite{jensen1906fonctions}. We will separately discuss the components in RHS of Eq.~\ref{eq:a_e} below. First, by importance sampling~\cite{kloek1978bayesian}, the outer expectation over data distribution would be:
\begin{equation}
\begin{aligned}
    \mathbb{E}_{(\bm{x},y)\sim p_{test}(\bm{x},y)}\psi(\bm{x},y)&=\int_{\mathcal{D}}\psi(\bm{x},y)p_{test}(\bm{x},y)d(\bm{x},y) \\
    &=\int_{\mathcal{D}}\frac{p_{test}(\bm{x},y)}{p_{train}(\bm{x},y)}\psi(\bm{x},y)p_{train}(\bm{x},y)d(\bm{x},y) \\
    &=\mathbb{E}_{(\bm{x},y)\sim p_{train}(\bm{x},y)}\frac{p_{test}(\bm{x},y)}{p_{train}(\bm{x},y)}\psi(\bm{x},y),\label{eq:out}
\end{aligned}
\end{equation}
where $\psi(\bm{x},y)$ denotes any expression with respect to $(\bm{x},y)$. Second, for the part inside the integral, we have:
\begin{equation}
\begin{aligned}
    &\int_{\Theta}q(\bm{\theta})\log{\left[g(\bm{\theta},d(\bm{x}))\frac{p(\bm{\theta}|\mathcal{D}_{test})}{q(\bm{\theta})}\right]}d\bm{\theta} \\
    &=\int_{\Theta}q(\bm{\theta})\log{\left[g(\bm{\theta},d(\bm{x}))\cdot\frac{p(\bm{\theta})}{q(\bm{\theta})}\cdot\frac{p(\mathcal{Y}_{test}|\mathcal{X}_{test},\bm{\theta})}{p(\mathcal{Y}_{test}|\mathcal{X}_{test})}\right]}d\bm{\theta} \\
    &=\int_{\Theta}q(\bm{\theta})\left[\log{g(\bm{\theta},d(\bm{x}))}-\log{\frac{q(\bm{\theta})}{p(\bm{\theta})}}+\log{\prod_tp(y_t|\bm{x}_t,\bm{\theta})}-\log{p(\mathcal{Y}_{test}|\mathcal{X}_{test})}\right]d\bm{\theta} \\
    &=\mathbb{E}_{\bm{\theta}\sim q(\bm{\theta})}\log{g(\bm{\theta},d(\bm{x}))}-KL(q(\bm{\theta})||p(\bm{\theta}))+\sum_t\mathbb{E}_{\bm{\theta}\sim q(\bm{\theta})}\log{p(y_t|\bm{x}_t,\bm{\theta})}-\log{p(\mathcal{Y}_{test}|\mathcal{X}_{test})} \\
    &=\mathbb{E}_{\bm{\theta}\sim q(\bm{\theta})}\sum_{y'}u(y',d)\log{p(y'|\bm{x},\bm{\theta})}-KL(q(\bm{\theta})||p(\bm{\theta}))+\mathbb{E}_{(\bm{x},y)\sim p(\mathcal{D}_{test})}\mathbb{E}_{\bm{\theta}\sim q(\bm{\theta})}\log{p(y|\bm{x},\bm{\theta})}-\log{p(\mathcal{Y}_{test}|\mathcal{X}_{test})}.\label{eq:in}
\end{aligned}
\end{equation}
Third, combining Eq.~\ref{eq:out} and Eq.~\ref{eq:in}, we have:
\begin{equation}
    \log{G(d)}\geq\mathbb{E}_{(\bm{x},y)\sim p_{train}(\bm{x},y)}\mathbb{E}_{\bm{\theta}\sim q(\bm{\theta})}\frac{p_{test}(\bm{x},y)}{p_{train}(\bm{x},y)}\left[\log{p(y|\bm{x},\bm{\theta})}+\sum_{y'}u(y',d)\log{p(y'|\bm{x},\bm{\theta})}\right]-KL(q(\bm{\theta})||p(\bm{\theta}))+C,
\end{equation}
where $C=-\log{p(\mathcal{Y}_{test}|\mathcal{X}_{test})}$, as desired. \qed

\section{Relationship between $q(\bm{\theta})$ and $p(\bm{\theta}|\mathcal{D}_{test})$\label{appendix:kl}}
\emph{Proof.} We show the relationship between $q(\bm{\theta})$ and $p(\bm{\theta}|\mathcal{D}_{test})$ by computing the KL divergence between them:
\begin{equation}
\begin{aligned}
    KL(q(\bm{\theta})||p(\bm{\theta}|\mathcal{D}_{test}))&=\int_\Theta q(\bm{\theta})\log{\frac{q(\bm{\theta})}{p(\bm{\theta}|\mathcal{X}_{test},\mathcal{Y}_{test})}}d\bm{\theta} \\
    &=\int_\Theta q(\bm{\theta})\log{\frac{q(\bm{\theta})p(\mathcal{Y}_{test}|\mathcal{X}_{test})}{p(\bm{\theta})p(\mathcal{Y}_{test}|\mathcal{X}_{test},\bm{\theta})}}d\bm{\theta} \\
    &=\int_\Theta q(\bm{\theta})\left[\log{\frac{q(\bm{\theta})}{p(\bm{\theta})}}-\sum_t\log{p(y_t|\bm{x}_t,\bm{\theta})}+\log{p(\mathcal{Y}_{test}|\mathcal{X}_{test})}\right]d\bm{\theta} \\
    &=KL(q(\bm{\theta})||p(\bm{\theta}))-\mathbb{E}_{(\bm{x},y)\sim p_{test}(\bm{x},y)}\mathbb{E}_{\bm{\theta}\sim q(\bm{\theta})}\log{p(y|\bm{x},\bm{\theta})}+\log{p(\mathcal{Y}_{test}|\mathcal{X}_{test})} \\
    &\overset{\text{(c)}}{=}KL(q(\bm{\theta})||p(\bm{\theta}))-\mathbb{E}_{(\bm{x},y)\sim p_{train}(\bm{x},y)}\mathbb{E}_{\bm{\theta}\sim q(\bm{\theta})}\frac{p_{test}(\bm{x},y)}{p_{train}(\bm{x},y)}\log{p(y|\bm{x},\bm{\theta})}-C.
\end{aligned}
\end{equation}
Here, (c) is by importance sampling (see Eq.~\ref{eq:out}) and $C=-\log{p(\mathcal{Y}_{test}|\mathcal{X}_{test})}$. \qed

\section{Implementation Details\label{appendix:id}}
Due to practical reasons, we slightly modify the training objective of our framework in experiments.
\begin{table}[ht]\setlength\tabcolsep{5pt}
\centering
\caption{hyper-parameter configurations.}
\label{tab:hyper}
\vskip 0.15in
\begin{tabular}{c|ccccccccc}
\hline
\textsc{Dataset} & \begin{tabular}[c]{@{}c@{}}\textsc{Base}\\ \textsc{model}\end{tabular} & \textsc{Optimizer} & \begin{tabular}[c]{@{}c@{}}\textsc{Batch}\\ \textsc{size}\end{tabular} & \begin{tabular}[c]{@{}c@{}}\textsc{Learning}\\ \textsc{rate}\end{tabular} & \begin{tabular}[c]{@{}c@{}}\textsc{Training}\\ \textsc{epochs}\end{tabular} & \begin{tabular}[c]{@{}c@{}}\textsc{Discrepancy}\\ \textsc{ratio}\end{tabular} & $\lambda$ & $\tau$ & $\alpha$ \\ \hline
CIFAR-10-LT      & ResNet32                                                               & SGD                & 128                                                                    & 0.1                                                                       & 200                                                                         & linear                                                                        & 5e-4      & 40     & 0.002    \\
CIFAR-100-LT     & ResNet32                                                               & SGD                & 128                                                                    & 0.1                                                                       & 200                                                                         & linear                                                                        & 5e-4      & 40     & 0.3      \\
ImageNet-LT      & ResNet50                                                               & SGD                & 256                                                                    & 0.1                                                                       & 100                                                                         & linear                                                                        & 2e-4      & 20     & 50       \\ \hline
\end{tabular}
\end{table}

\paragraph{Repulsive force.} We find in experiments that although applying repulsive force can promote the diversity of particles, it will certainly disturb the fine-tuning stage in training, which consequently results in sub-optimal performances by the end of training. To address this issue, we apply an annealing weight to the repulsive force to reduce its effect as the training proceeds:
\begin{equation}
    \exp\{-epoch/\tau\}\cdot\frac{1}{2}\sum_j\log{(\overline{\bm{\theta}^2}-\overline{\bm{\theta}}^2)_j},
\end{equation}
where $\tau$ is a stride factor which controls the decay of annealing weight. With the annealing weight, the repulsive force will push particles away at the beginning of training, and gradually become negligible at the end of training.

\paragraph{Tail-sensitive utility.} Although the tail-sensitive utility matrix in Fig.~\ref{fig:u} is designed to address the problem of classifying too many tailed samples into head classes, it will also affect the accuracy of classification task. Therefore, the utility term in Eq.~\ref{eq:obj} needs re-scaling so that its negative effect on the accuracy is controllable:
\begin{equation}
    \log{p(y|\bm{x},\bm{\theta})}+\frac{1}{\alpha}\cdot\sum_{y'}u(y',d)\log{p(y'|\bm{x},\bm{\theta})},
\end{equation}
where $\alpha$ is the scaling factor. We can adjust the value of $\alpha$ to carefully control the effect of the tail-sensitive utility term, which will bring to us significant improvement on the False Head Rate with acceptable accuracy drop.

\paragraph{Computational cost.} The model architecture follows RIDE~\cite{wang2020long} and TLC~\cite{li2022trustworthy}, in which the first few layers in neural networks are shared among all particles. Therefore, the computational cost of LBD is comparable to previous ensemble models. Besides, compared with gradient-flow-based BNN like \cite{d2021repulsive}, which typically uses 20 particles, our model is far more efficient with no more than 5 particles.

Other settings and hyper-parameters are concluded in Table~\ref{tab:hyper}. The optimal values of those hyper-parameters are determined by grid search.



\end{document}